%% file: main.tex
\documentclass[sigconf]{acmart}
\usepackage[utf8]{inputenc} %
\usepackage[T1]{fontenc}    %
\usepackage{hyperref}       %
\usepackage{url}            %
\usepackage{booktabs}       %
\usepackage{amsmath,amsfonts}
\usepackage{nicefrac}       %
\usepackage{microtype}      %
\usepackage{xcolor}         %
\usepackage{multicol}
\usepackage{multirow}
\usepackage{graphicx}
\usepackage{makecell}
\usepackage{wrapfig,lipsum,booktabs}
\usepackage{enumitem}
\usepackage{ulem}
\usepackage{nth}
\usepackage{bbm}
\usepackage{mdwlist}
\usepackage[font=small]{subcaption}
\usepackage{comment}
\definecolor{darkred}{RGB}{192, 0, 0}
\DeclareMathAlphabet\mathbfcal{OMS}{cmsy}{b}{n}

\usepackage{yfonts}
\usepackage{microtype}

\newcommand{\ignore}[1]{}

\newcommand{\new}[1]{\textcolor{blue}{#1}}
\renewcommand{\new}[1]{{#1}}

\newtheorem{problem}{Task}

\newcommand\mypar[1]{\noindent \textbf{#1}}
\renewcommand{\paragraph}[1]{\noindent\textbf{#1.}}

\title{Leveraging the Graph Structure of Neural Network Training Dynamics}

\author{Fatemeh Vahedian}
\affiliation{
 \institution{University of Michigan}
 \country{Ann Arbor, USA}
}
\email{vfatemeh@umich.edu}
\author{Ruiyu Li}
\affiliation{
 \institution{University of Michigan}
 \country{Ann Arbor, USA}
 }
 \email{ruiyuli@umich.edu}
\author{Puja Trivedi}
\affiliation{
 \institution{University of Michigan}
 \country{Ann Arbor, USA}
}
\email{pujat@umich.edu} 
\author{Di Jin}
\affiliation{
  \institution{University of Michigan}
 \country{Ann Arbor, USA}
}
\email{dijin@umich.edu}
\author{Danai Koutra}
\affiliation{
  \institution{University of Michigan}
  \country{Ann Arbor, USA}
}
\email{dkoutra@umich.edu}

\renewcommand{\shortauthors}{Vahedian, Li, Trivedi, Jin, Koutra.}

\copyrightyear{2022}
\acmYear{2022}
\setcopyright{acmcopyright}\acmConference[CIKM '22]{Proceedings of the 31st ACM International Conference on Information and Knowledge Management}{October 17--21, 2022}{Atlanta, GA, USA}
\acmBooktitle{Proceedings of the 31st ACM International Conference on Information and Knowledge Management (CIKM '22), October 17--21, 2022, Atlanta, GA, USA}
\acmPrice{15.00}
 \acmDOI{10.1145/3511808.3557628}
\acmISBN{978-1-4503-9236-5/22/10}

\input{dfn}
\begin{document}

\begin{abstract}
  \input{PAGES_v3/000abstract}
\end{abstract}
\maketitle

\section{Introduction}
\input{PAGES_v3/010introduction}

\section{Proposed NN Graph Representation}
\input{PAGES_v3/020proposedrep.tex}

\section{Proposed Temporal Framework}
\input{PAGES_v3/030methods.tex}

\section{Empirical Analysis}\label{exp}
\input{PAGES_v3/040experiments}
\section{Conclusion}
\input{PAGES_v3/050conclusion.tex}

\newpage
\section*{Acknowledgements}
This material is based upon work supported by 
the NSF under Grant No. IIS 1845491, and Amazon and Facebook faculty awards.
Any opinions, findings, conclusions or recommendations 
expressed in this material are those of the {authors} and do not necessarily reflect the views of the National Science Foundation or other funding parties.
\balance 
\normalem
\bibliographystyle{iclr2022_conference}
\bibliography{all}

\end{document}

%% file: dfn.tex
\definecolor{darkgreen}{RGB}{0,153,0}

\newlist{steps}{enumerate}{1}
\setlist[steps, 1]{label = Step \arabic*:}

\newcommand{\graph}{\mathbf{G}}

\newcommand{\tenK}{\mathbfcal{K}}

\newcommand{\matW}{\mathbf{W}}

\newcommand{\vecf}{\mathbf{s}}

\newcommand{\setN}{\mathcal{N}}

\newcommand{\tempG}{\mathcal{G}}

\newcolumntype{H}{>{\setbox0=\hbox\bgroup}c<{\egroup}@{}}

\makeatletter
\def\adl@drawiv#1#2#3{%
        \hskip.5\tabcolsep
        \xleaders#3{#2.5\@tempdimb #1{1}#2.5\@tempdimb}%
                #2\z@ plus1fil minus1fil\relax
        \hskip.5\tabcolsep}
\newcommand{\cdashlinelr}[1]{%
  \noalign{\vskip\aboverulesep
           \global\let\@dashdrawstore\adl@draw
           \global\let\adl@draw\adl@drawiv}
  \cdashline{#1}
  \noalign{\global\let\adl@draw\@dashdrawstore
           \vskip\belowrulesep}}
\makeatother

%% file: PAGES_v3/000abstract.tex
Understanding the training dynamics of deep neural networks (DNNs) is important  as it can lead to improved training efficiency and task performance. Recent works have demonstrated that representing the wirings of neurons in feedforward DNNs as graphs is an effective strategy for understanding how architectural choices can affect performance. However, these approaches fail to model training dynamics since a single, \textit{static} graph cannot capture how DNNs change over the course of training. Thus, in this work, we propose a compact, expressive \textit{temporal} graph framework that effectively captures the dynamics of many workhorse architectures in computer vision. Specifically, it extracts an informative summary of graph properties (e.g., eigenvector centrality) over a sequence of DNN graphs obtained during training. 
We demonstrate that our framework captures useful dynamics by accurately predicting trained, task performance when using a summary over early training epochs (<5) across four different architectures and two image datasets. Moreover, by using a novel, highly-scalable DNN graph representation, we also show that the proposed framework captures generalizable dynamics as summaries extracted from smaller-width networks are effective when evaluated on larger widths\footnote{Code: \href{https://github.com/pindapuj/NN_Code.git}{\texttt{https://github.com/pindapuj/NN\_Code.git}}}. 

{
\small

\ccsdesc[500]{Computing methodologies~Machine learning}
}

%% file: PAGES_v3/010introduction.tex
The impressive success of deep neural networks (DNNs) in machine learning tasks across a variety of domains \cite{8578843,sutskever2014sequence,bahdanau2014neural,cao2020ensemble,li2019deep} has led to considerable interest in understanding how the interplay between the training process, model architecture, hyper-parameters and other factors influences task performance \citep{raghu2017svcca,8397411,ioffe2015batch,he2016deep}. Key to this endeavor is a representation or framework for studying DNNs that is amenable to jointly analyzing these factors \citep{jacot2018neural}. 
Recognizing the intuitive relationship between the wiring of neurons in DNNs and graphs, recent works \citep{you2020graph,filan2021clusterability,Rieck19a} have proposed \textit{graph} representations of DNNs and sought to leverage network science to understand how properties of the resulting representation are related to performance or behavior. 
Such graph representations are intuitive, provide a unified language (e.g., network science) for discussing properties of different architectures and have been shown to be effective at understanding static properties of DNNs. 

However, existing graph representations are unable to capture how DNNs change throughout training and, therefore, cannot be effectively used to understand DNN training dynamics. Indeed, capturing dynamics is difficult as DNN graphs must not only be generated over time but also be expressive enough to meaningfully model how DNN parameters evolve. Existing graph representations are either unable to effectively scale for temporal settings due to untenable memory requirements~\cite{Rieck19a} or are not expressive enough for a dynamic setting~\cite{you2020graph}. Therefore, we introduce a new graph representation and a corresponding framework that is able to model the training dynamics of many workhorse architectures in computer vision, while also preserving the benefits of a graph representation. 

\vspace{0.13cm}
\mypar{Present Work.} 
In this paper, we propose a compact, graph representation and corresponding temporal framework for better understanding DNN training dynamics. Specifically, we first construct a series of graphs over the course of training, and then create informative summaries of graph properties (e.g., weighted degree, eigenvector centrality) to understand how the training induces structural changes in the DNN graph. To demonstrate the utility of the proposed framework, we use the temporal summaries extracted from a few early training epochs (<5) as effective features in the challenging task of predicting the final performance of fully-trained DNNs. Notably, reliably predicting performance is practically useful %
for early stopping~\citep{yu2020hyper}. Our main contributions are: 

\setlist{leftmargin=*}
\begin{itemize}
    \item \textbf{Compact, expressive graph representation of DNN:} We introduce a graph representation for convolutional layers that is significantly more compact than existing graph representations~\citep{Rieck19a}, enabling use in the analysis of the training dynamics.
    
    \item \textbf{Graph framework for NN performance prediction:} We propose a temporal framework that summarizes sequences of DNN graphs to effectively model structural changes during training.
    
    \item \textbf{Extensive empirical analysis:} 
    Across several DNN architectures (AlexNet, VGG, LeNet,  ResNet) and two image datasets, we verify the utility our framework in the challenging task of final performance prediction from early training epochs. {Indeed, with temporal summaries over less than 5 epochs of training, our framework achieves classification accuracy of 90\%.} We also show that it captures generalizable dynamics by extracting summaries from smaller-width networks and predicting on large widths.   
\end{itemize}

%% file: PAGES_v3/020proposedrep.tex
\label{sub:roll}
We now introduce our proposed compact representation of convolutional layers that can effectively and efficiently capture training dynamics. We begin by  discussing existing graph representations. 

\vspace{0.13cm}
\mypar{DNNs as Graphs: Related Work \& Limitations.} 
Existing works primarily focus on representing DNNs which contain only fully connected (fc) and convolutional (conv) layers as these are the key components of many popular vision architectures (LeNet~\citep{lenet}, VGG~\citep{vgg} and ResNet~\citep{he2016deep}). 
\citeauthor{you2020graph}~\cite{you2020graph} propose representing DNNs as relational graphs where edges correspond to message passing between layers (nodes), and show that the graph clustering coefficient and average path length  can identify a ``sweet spot'' for task performance. Rieck et al.~\cite{Rieck19a} leverage weighted, stratified DNN graphs, where neurons correspond to nodes, and weights correspond to edges. They introduce a corresponding complexity measure which is empirically well-aligned with best training practices. Filan et al.~\cite{filan2021clusterability} focus on weighted graph representation for MLPs only, where each neuron---including 
the input/output layers---corresponds to a node, and two neurons are linked if they appear in consecutive layers. 
They find that DNNs are ``surprisingly modular.'' 

While these representations provide interesting insights into DNNs, they are not suitable for understanding dynamics. 

\begin{figure}[t!]
\captionsetup[subfigure]{justification=centering}
      \centering
       \includegraphics[width=\columnwidth]{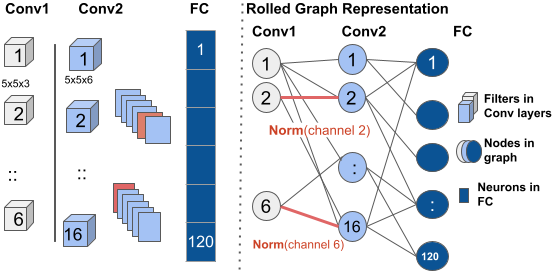}
  \caption{\textit{Rolled graph representation example}: The resultant graph is a tri-partite graph with three node types (Conv1, Conv2, FC).  
  }
\label{fig:roll_graph}
\end{figure}

Unweighted graphs cannot represent how convolutional filters or  neurons change over training. 
Weighted edges in graph representations of fc layers~\cite{filan2021clusterability,Rieck19a} 
suggest a mechanism for capturing dynamics. However, \citeauthor{Rieck19a} ``unroll'' the convolution operator so each position in operation maps to a node in the graph and each multiplication maps to a weighted edge. Besides destroying the semantics of filters, unrolling leads to an explosion of nodes/edges and  untenable memory requirements, as understanding dynamics requires creating DNN graphs over some epochs (time steps).\label{sub:nn}
Motivated by the inefficiencies and promise of the unrolled graph representation~\cite{Rieck19a}, we propose a compact, ``rolled'' representation for conv layers (Fig. \ref{fig:roll_graph}), which is not only more interpretable than the unrolled model, but also highly effective despite its low footprint  {(\S~\ref{exp})}. 

\vspace{0.13cm}
\noindent \textbf{Rolled Graph Representation of Conv Layers.} Let tensor, $\tenK_i$, be a kernel in layer $i$ with $f_i$ filters, each with $ c_i$ channels and dimensions  $h_i \times w_i $. We use bracket notation to index into the kernel: for example, $\tenK_i [ l,:,:,:]$ indexes the $l^\text{th}$ filter of kernel $\tenK_i$.  
We create $f_i$ nodes representing each 
filter $\{v^{(i)}_{1},v^{(i)}_{2},...v^{(i)}_{f_i}\}$, 
where node features are defined as the corresponding biases in that layer. 
Then, $\tenK_j$ is the next convolutional layer, defined analogously.
While edges between neurons in fc layers are directly defined through neuron weights, each kernel contains multiple weights. To extract a weighted edge, we take the norm over each kernel's channels.
Formally, the edge between node $v^{(i)}_{k}$ representing the $k^{th}$ filter in layer $i$ (i.e., $\tenK_i[k,:,:,:]$) and node/filter $v^{(j)}_l$ in layer $j=i+1$ (i.e. $\tenK_j[l,:,:,:]$) has weight 
$w_{v^{(i)}_k,v^{(j)}_l} = \text{norm}(\tenK_j[l,k,:,:])$,  which is the norm of the $k^\text{th}$ channel of the  $l^\text{th}$filter in the $j^\text{th}$  layer.
While alternative edge weight configurations can be supported, we focus on the filter norm as other studies, including those on pruning NNs \citep{li2016pruning}, have demonstrated that this value has strong correlation with filter importance.
As shown in Fig. \ref{fig:roll_graph}, in the case of two conv layers (i.e., w/o the fc layer), the resultant graph is an attributed bipartite graph with $f
  _i+f_j$ nodes and  $f_i \times f_j$ edges, where
node attributes include flattened weight vectors or filter maps. 
Other information such as average gradients can also be used as node features.

%% file: PAGES_v3/030methods.tex
\label{sec:method}

Successful performance prediction based on only a few epochs could be used for early stopping~\citep{yu2020hyper}, and thus, more efficient NN training. 
We show the utility of our proposed  graph representation,
by investigating the relationship between the temporal graph structure of NNs in early training stages and NN performance in downstream tasks. Formally: 

\begin{problem}[NN Performance Prediction]
Let $\setN=\{N_{tr_1}, \ldots,N_{tr_n}\}$ be a training set of $n$ NNs trained for $T$ epochs and 
$\mathcal{A}=\{ \alpha_{1}, \alpha_{2},\ldots,\alpha_{n} \}$
their corresponding downstream task accuracies (e.g., for image classification).
We seek to predict the accuracy $\alpha_{tst}$
of a new instance $N_{tst}$  trained for a very small number of $t \ll T$ epochs 
by using  $t$ epochs for the trained NNs in $\setN$. 
\label{def:problem-statement}
\end{problem}

\begin{figure}[t]
    \centering
    \includegraphics [width=\columnwidth]{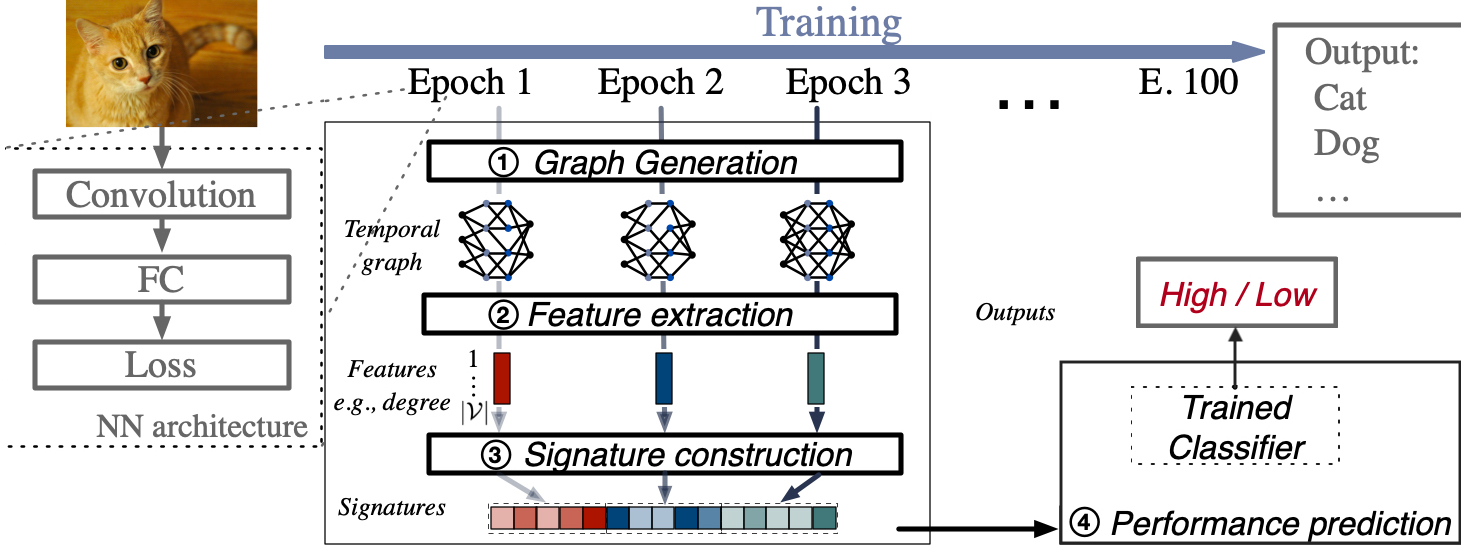}
    \caption{Our proposed framework for predicting NN performance in a downstream image classification task, shown for one test instance. 
    }
    \label{fig:frm}
\end{figure}

\label{sub:frame}
Building on temporal graph mining and summarization~\cite{Holme15,RozenshteinG19,LiuSDK18-survey,BelthZK20},  
to tackle this problem, we introduce a temporal graph-based framework (Fig.~\ref{fig:frm}), which consists of four steps: 

\vspace{0.13cm}
\noindent \textbf{Step (S1) Graph generation.} 
This step involves converting the training process of each input NN $N_{tr_i} \in \setN$ into a time-evolving graph where each static snapshot corresponds to a different epoch (time step). 
The output is a set of $n$ temporal graphs 
$\{\tempG_{tr_1}, ...,\tempG_{tr_n}\}$, where $\tempG_{tr_i} = \{\graph^{1}_i, ...,\graph^{t}_i\}$ 
corresponds to the $i^{th}$ original NN in $\setN$. 

\vspace{0.13cm}
\noindent {\textbf{Step (S2) Feature extraction.}}
Next, the goal is to capture the structural dynamics of the NN training process. We aim to select graph measures that can capture changes during the training process, take into account the edge weight of graphs and can be calculated efficiently.
In order to do that in an interpretable way, we extract two well-known node centralities from each snapshot of each generated time-evolving graph $\tempG_i$: weighted degree centrality and eigenvector centrality~\citep{bonacich1972factoring}. 
The weighted degree is a simple function of the learnable weight matrix $\matW$ during the training phase of NN, therefore it gives us insights into the training dynamics at the node/neuron/filter level. 
The eigenvector centrality is an extension of the degree centrality, which captures the highly influential nodes,
and has been successfully used in neuroscience to capture the dynamic changes of real neural networks (or connectomes)~\citep{lohmann2010eigenvector}. Eigenvector centrality can be used to capture importance and connectivity of filters/neurons (i.e., the nodes in our graph representation). 
\new{It has also been used for detecting communities \citep{newman2006finding} or clusters \citep{wu2013follow}, and thus provides structural information about the clusterability of the NN, which is  complementary to that provided by the simpler and more efficient-to-compute degree centrality. }
\new{Our choice of features is guided by the inherent $k$-partite structure of our proposed graph representation, 
which cannot be captured well by  other commonly-used graph features (e.g., clustering-based features like triangles, transitivity, clustering coefficient are  0).}

\vspace{0.13cm}
\noindent \textbf{Step (S3) Graph signature construction.} 
\new{In order to be able to compare DNN graphs (with different number of nodes and edges)}~\cite{KoutraEF14,KoutraF17}, we summarize the structural changes in the generated temporal graphs at the \textit{graph} level (rather than the \textit{node} level, as in (S2)), and 
construct a statistical summary of 
the extracted node centralities (signature) per time-evolving graph $\tempG_i$.
For each snapshot $\graph_i^{(\tau)}$ of $\tempG_i$,
we create a signature vector using five node feature aggregators, which were introduced in \citep{BerlingerioKEF13} for graph similarity:
\textit{median, mean, standard deviation, skewness, and kurtosis}, 
where all but the median are moments of the corresponding  distribution. 
Thus, $\graph_i^{(\tau)}$ is mapped to a (static) signature vector $s_i^{(\tau)} \in \mathbb{R}^5$, representing  
the statistical summary of its node features (i.e., degree or eigenvector centrality) at time $\tau$. 
To put more emphasis on the most recent timesteps, 
we can redefine the signature at time $\tau$ as the linear weighted average of the signatures up to that point, {\small $\vecf_i^{(\tau)} \leftarrow \frac{\sum_{j=1}^\tau j*\vecf^{(j)}}{\sum j} $},
or an exponential function of the previous signatures,
{\small $\vecf_i^{(\tau)} \leftarrow \alpha \vecf_i^{(\tau)}+ (1-\alpha)\vecf_i^{(\tau-1)} $}. 
To obtain the temporal signature of the evolving graph $\tempG_i$, we aggregate the (static) signatures up to timestamp/epoch $t,\vecf^t_{i}=\vecf^1_{i}\oplus ... \oplus \vecf^t_{i},$
where $\oplus$ denotes concatenation.

\vspace{0.13cm}
\noindent \textbf{Step (S4) Performance prediction.}
For the performance prediction step, we consider a classification task:
We train a classifier (e.g., SVM, MLP) using 
the training graphs $\{\tempG_{tr_1}, \tempG_{tr_2}, ...,\tempG_{tr_n}\}$ 
represented by their temporal signatures 
$\{\vecf_{tr_1}^t, \vecf_{tr_2}^t, ...,\vecf_{tr_n}^t\}$,
and their corresponding accuracies $\mathcal{A}=\{ \alpha_{1}, \alpha_{2},\ldots,\alpha_{n} \}$ 
mapped to labels  $\mathcal{L}=\{ l_{1}, l_{2},\ldots, l_{n} \}$
(e.g., high/low accuracy) based on some threshold. 
Test NN instances are then classified by the trained model. 

%% file: PAGES_v3/040experiments.tex
\label{sec:analysis}
In this section, we empirically evaluate our framework in the NN performance prediction task.
We seek to answer three questions: 
\textbf{(Q1)} How effective is our framework when predicting the performance per DNN architecture? 
\textbf{(Q2)} Can our framework generalize to unseen architectures? 
\textbf{(Q3)} How efficient is it? 

\vspace{0.13cm}
\paragraph{Data} We investigate NN dynamics using two well-known image classification datasets, CIFAR-10 \citep{cf10} and ImageNet \citep{imagenet}. CIFAR-10 consists of 50K training images and 10K test images. For ImageNet, we use a sample that has 50K training images and 5K validation images used as the test set \citep{tiny}.

\vspace{0.13cm}
\noindent \textbf{Baselines.}
We compare our framework with variants of a graph-agnostic approach that was proposed in \cite{unterthiner2020predicting} to predict the accuracy of NNs directly from the learned weights. 
We consider three baseline methods using different feature vectors as input to our step (\textbf{S4}): 
\textbf{(1)}     $B_{W_l}$ leverages the flattened parameters (weights/kernels
and biases) of  the last layer, since, according to \cite{unterthiner2020predicting}, the parameters learned in the last dense layer are as informative as those from all the  layers for predicting NN performance\footnote{This observation was consistent with our experiments: the entire flattened vector across layers did not lead to higher prediction accuracy.};
 \textbf{(2)} $B_{\hat{W}}$ applies 7 aggregators to the flattened vector of the last layer: the mean, the variance, and $q^\text{th}$ percentiles for
$q \in \{0, 25, 50, 75, 100\}$;
\textbf{(3)} $B_{W^\prime}$ applies our 5 aggregators (from step (\textbf{S3}))  on the flattened vector of the last layer: median, mean, standard deviation, skewness, and kurtosis. We also compare our proposed rolled graph representation to \textbf{(4)} the previously proposed unrolled representation~\cite{Rieck19a}.
\begin{table}[t]
  \caption{Description of NNs: early stopping epoch range, accuracy, and accuracy threshold defining the  classes of the classification task. 
  } 
  \label{tab:NN}
  \centering
  \resizebox{\columnwidth}{!}{
  \setlength{\tabcolsep}{3pt}
  \begin{tabular}{lrrrrrcrrr}
    \toprule
    \multicolumn{1}{c}{} & \multicolumn{5}{c}{\bf  CIFAR-10 } && \multicolumn{3}{c}{\bf ImageNet}                  \\
    \cmidrule(r){2-6} \cmidrule(r){8-10}
         & \textbf{LeNet} & \textbf{AlexNet} & \textbf{VGG} & \textbf{ResNet-32} & \textbf{ResNet-44} && \textbf{LeNet} & \textbf{AlexNet} &  \textbf{ResNet-50}\\
    \midrule
    Early stop. &  11$\sim$50   & 30$\sim$50   & 45$\sim$50   & 16$\sim$120& 16$\sim$120 && 16$\sim$50  & 16$\sim$50 & 16$\sim$120 \\
    Acc.\ range     &9.4$\sim$73.8 &  5.5$\sim$82.4   &8.8$\sim$87.6  &8.4$\sim$90.0& 9.9$\sim$89.8 && 0.6$\sim$14.4    & 0.6$\sim$20.1 & 0.86$\sim$41.66 \\
    Acc.\ thres.     &40&  40  &40  & 40 & 40 &&  9  &  10 &  25 \\
    \bottomrule
  \end{tabular}
   }
\end{table}

\begin{figure*}[t]
\captionsetup[subfigure]{justification=centering}
  \centering
  \begin{subfigure}[t]{0.16\textwidth}
      \centering
      \includegraphics[width=.99\textwidth,trim={0 0 0 1cm}]{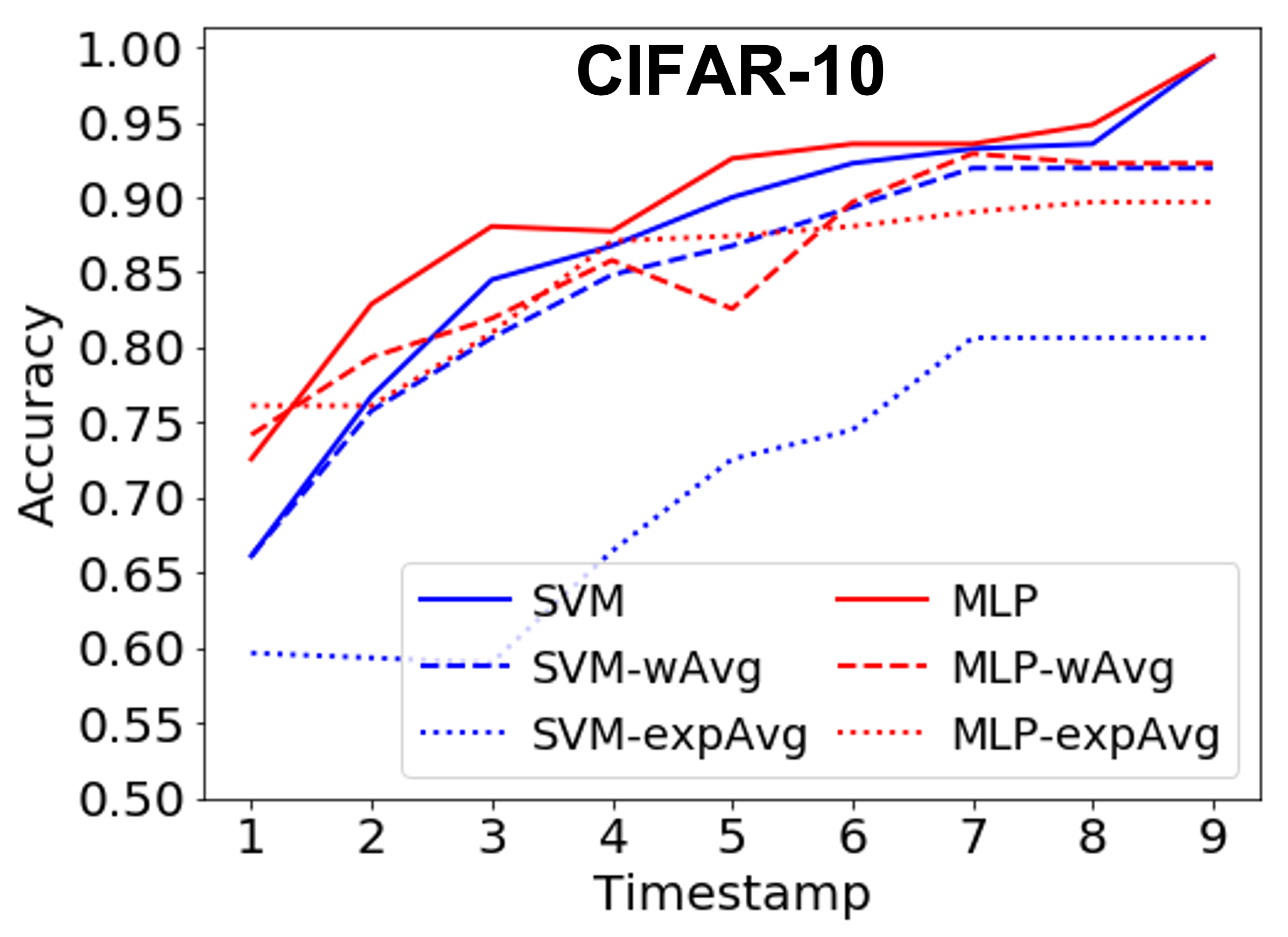}
      \caption{LeNet: deg}
  	\label{fig:cifar-roll-len-degree}
  \end{subfigure}
  \hfill 
     \begin{subfigure}[t]{0.16\textwidth}
      \centering
      \includegraphics[width=.99\textwidth,trim={0 0 0 1cm}]{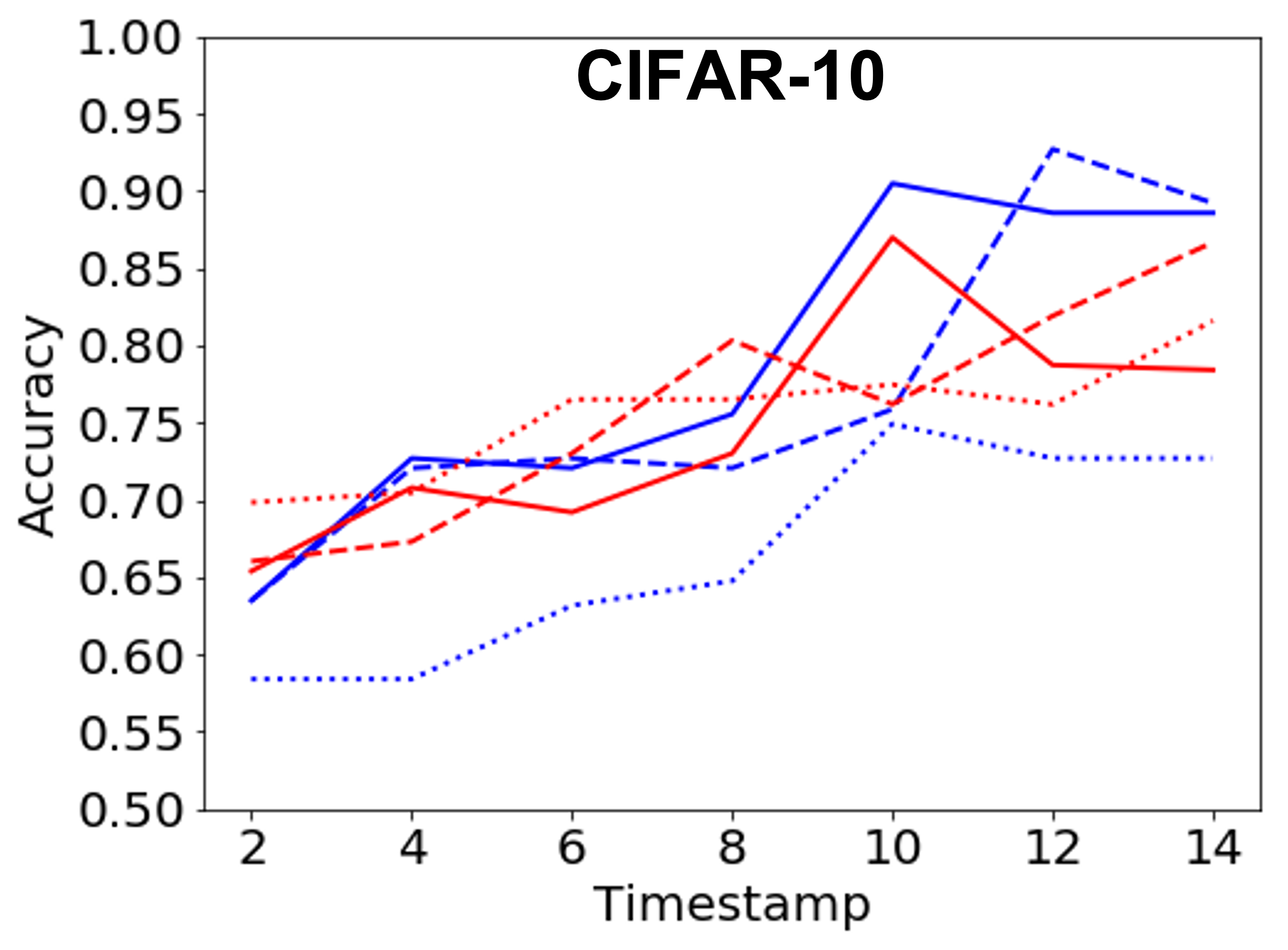}
      \caption{AlexNet: evec}
      \label{fig:cif-alex-roll-deg}
  \end{subfigure}
    \hfill
   \begin{subfigure}[t]{0.16\textwidth}
      \centering
      \includegraphics[width=.99\textwidth,trim={0 0 0 0.1cm}]{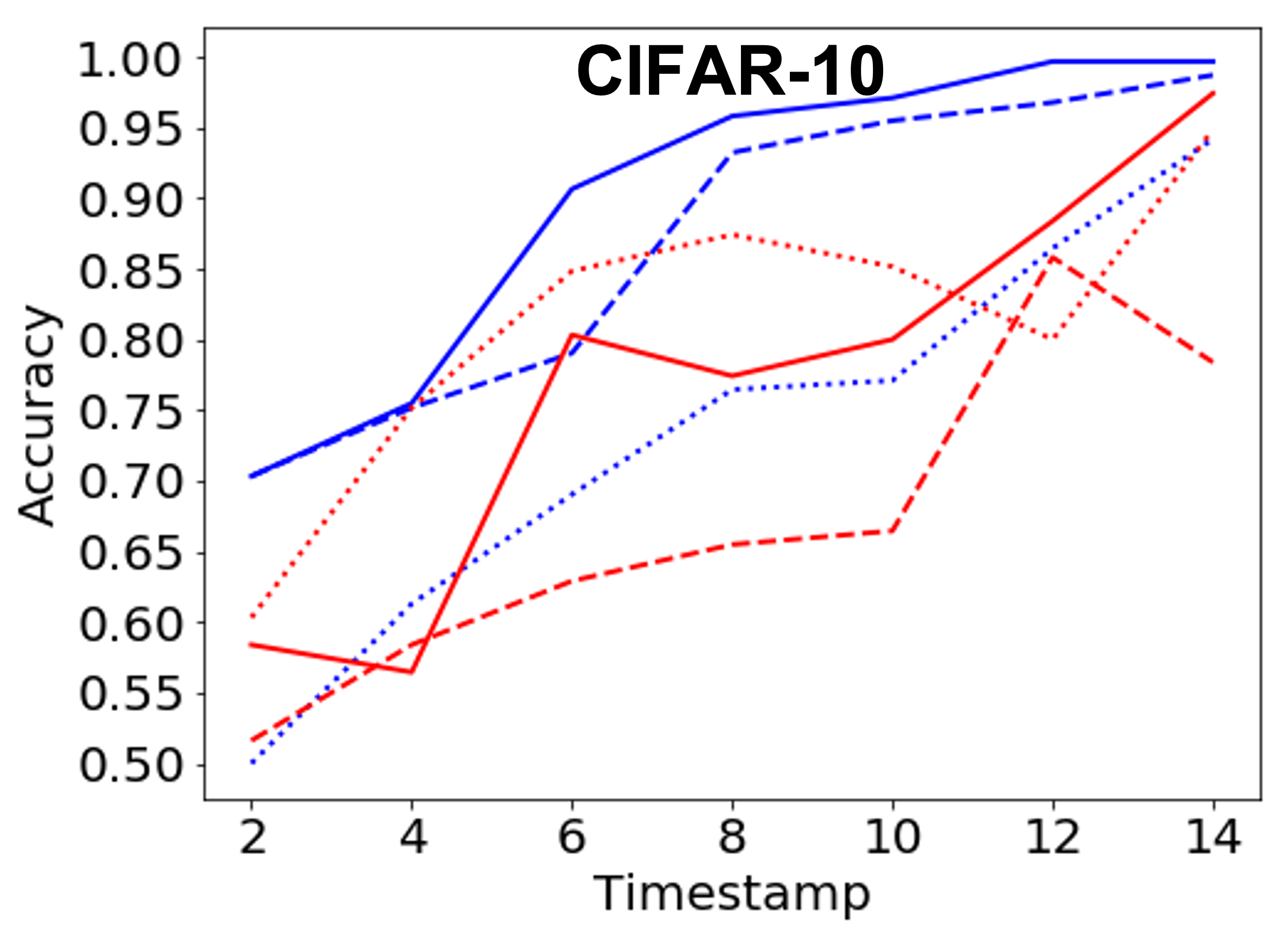}
      \caption{VGG: deg}
      \label{fig:cifar-vgg-deg}
  \end{subfigure}
  \hfill
   \begin{subfigure}[t]{0.16\textwidth}
      \centering
      \includegraphics[width=.99\textwidth,trim={0 0 0 0.1cm}]{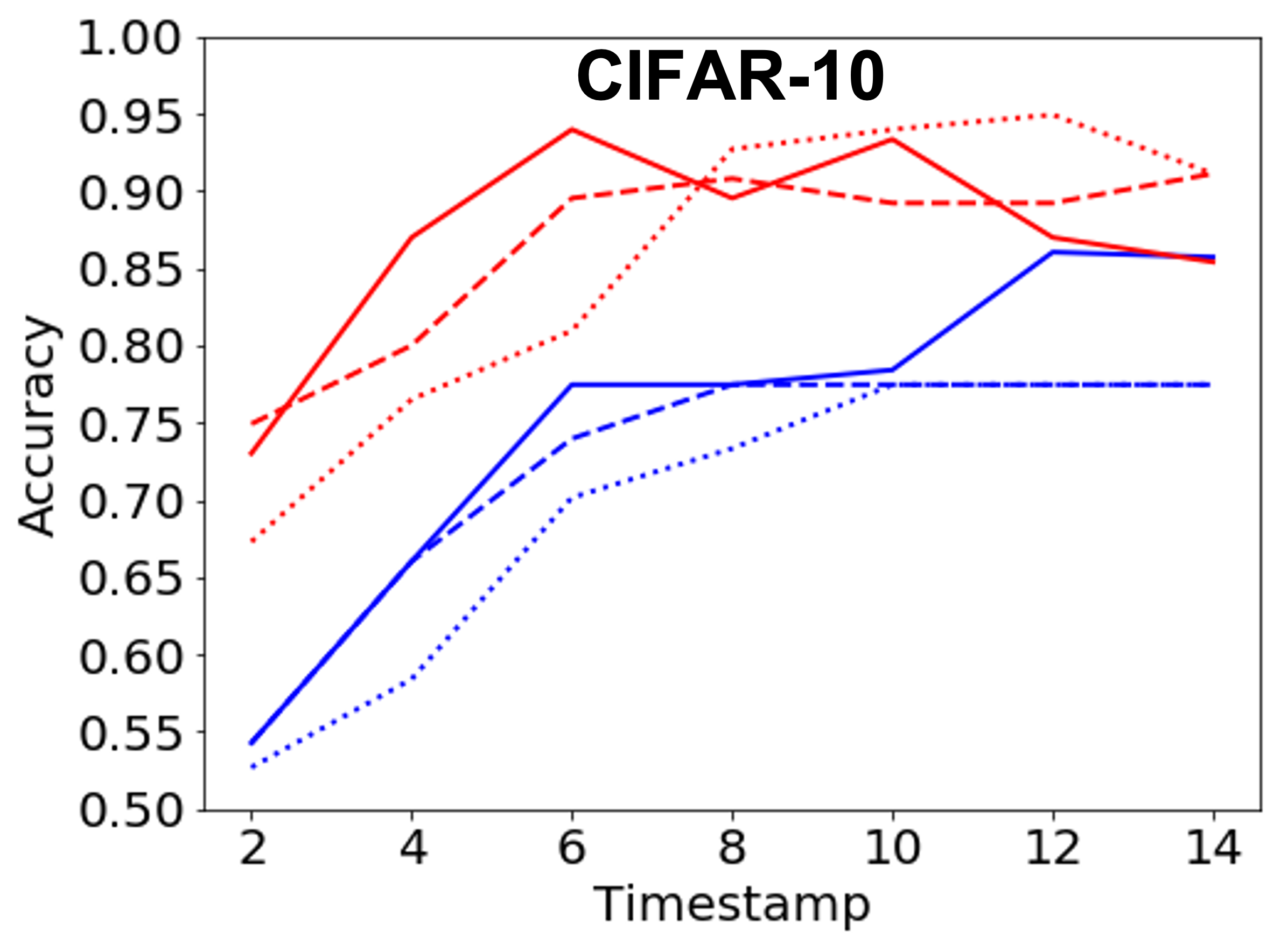}
      \caption{VGG: evec} 
      \label{fig:cifar-vgg-cen}
  \end{subfigure}
    \hfill
   \begin{subfigure}[t]{0.16\textwidth}
      \centering
      \includegraphics[width=.99\textwidth,trim={0 0 0 0.1cm}, clip]{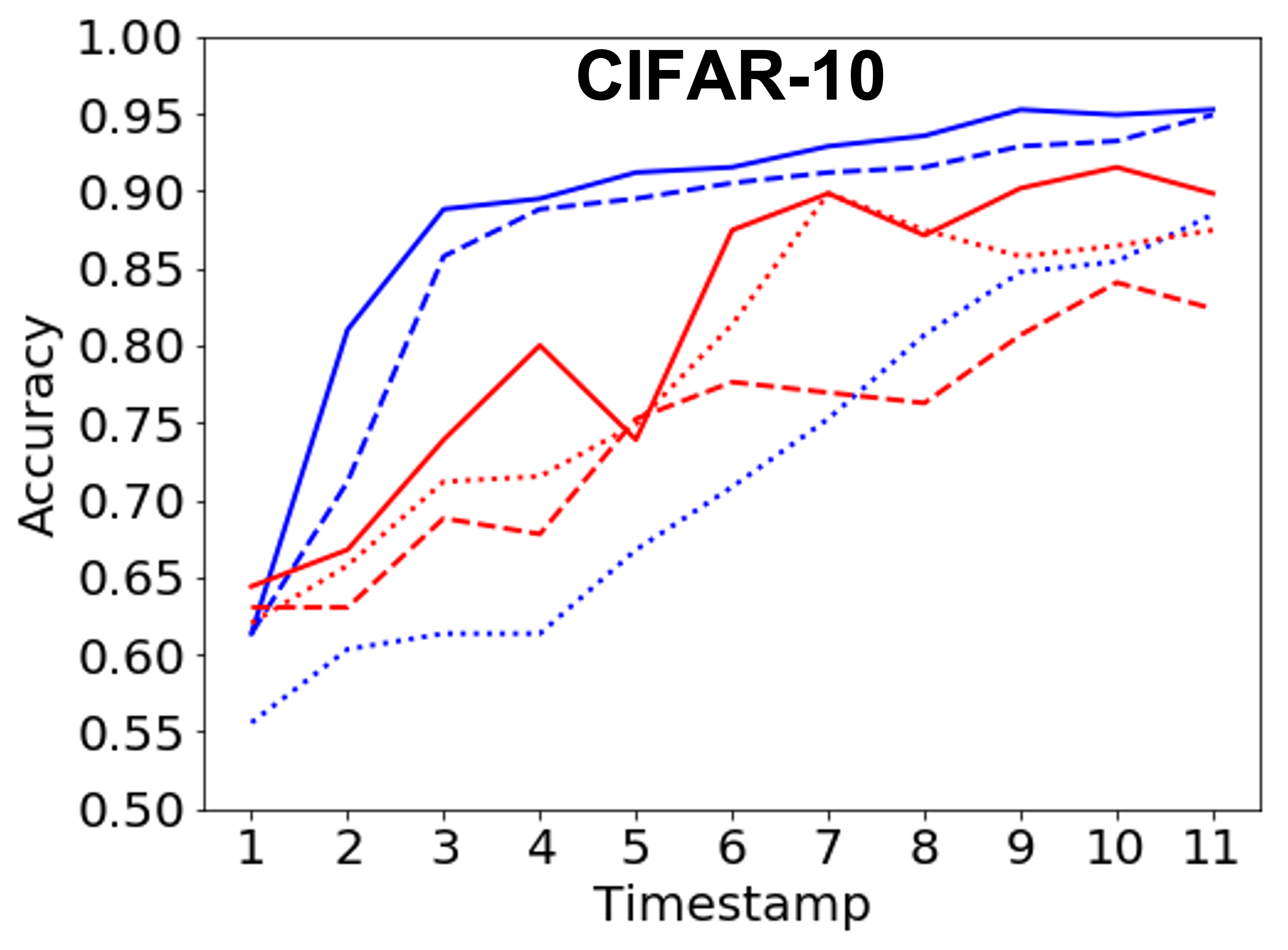}
      \caption{ResNet-44: deg }
      \label{fig:res44-deg}
  \end{subfigure}
  \hfill
   \begin{subfigure}[t]{0.16\textwidth}
      \centering
      \includegraphics[width=.99\textwidth,trim={0 0 0 0.1cm}, clip]{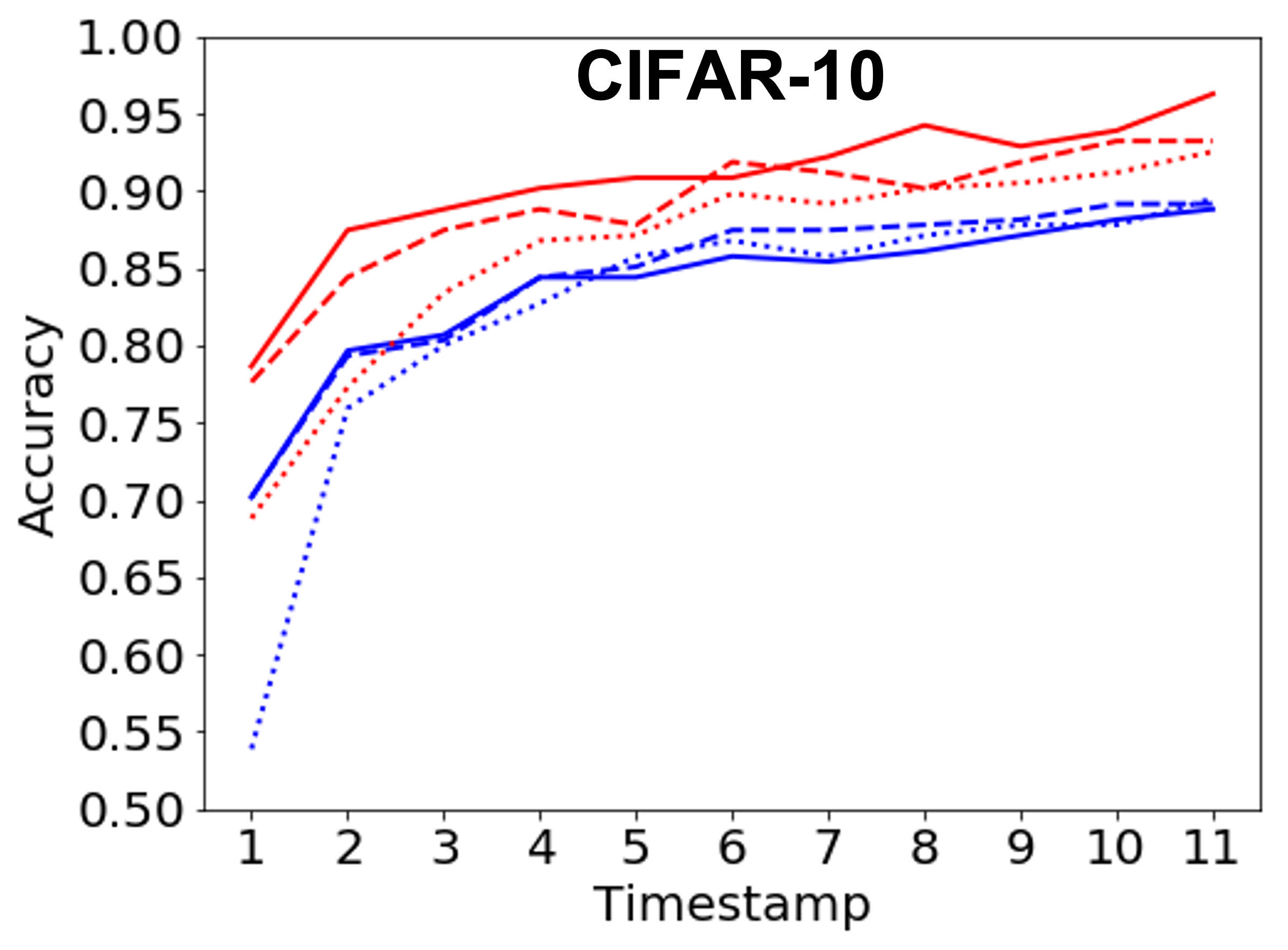}
      \caption{ResNet-44: evec } 
      \label{ig:res44-cent}
  \end{subfigure}
  
  \begin{subfigure}[t]{0.16\textwidth}
      \centering
      \includegraphics[width=\textwidth,trim={0 0 0 1cm}]{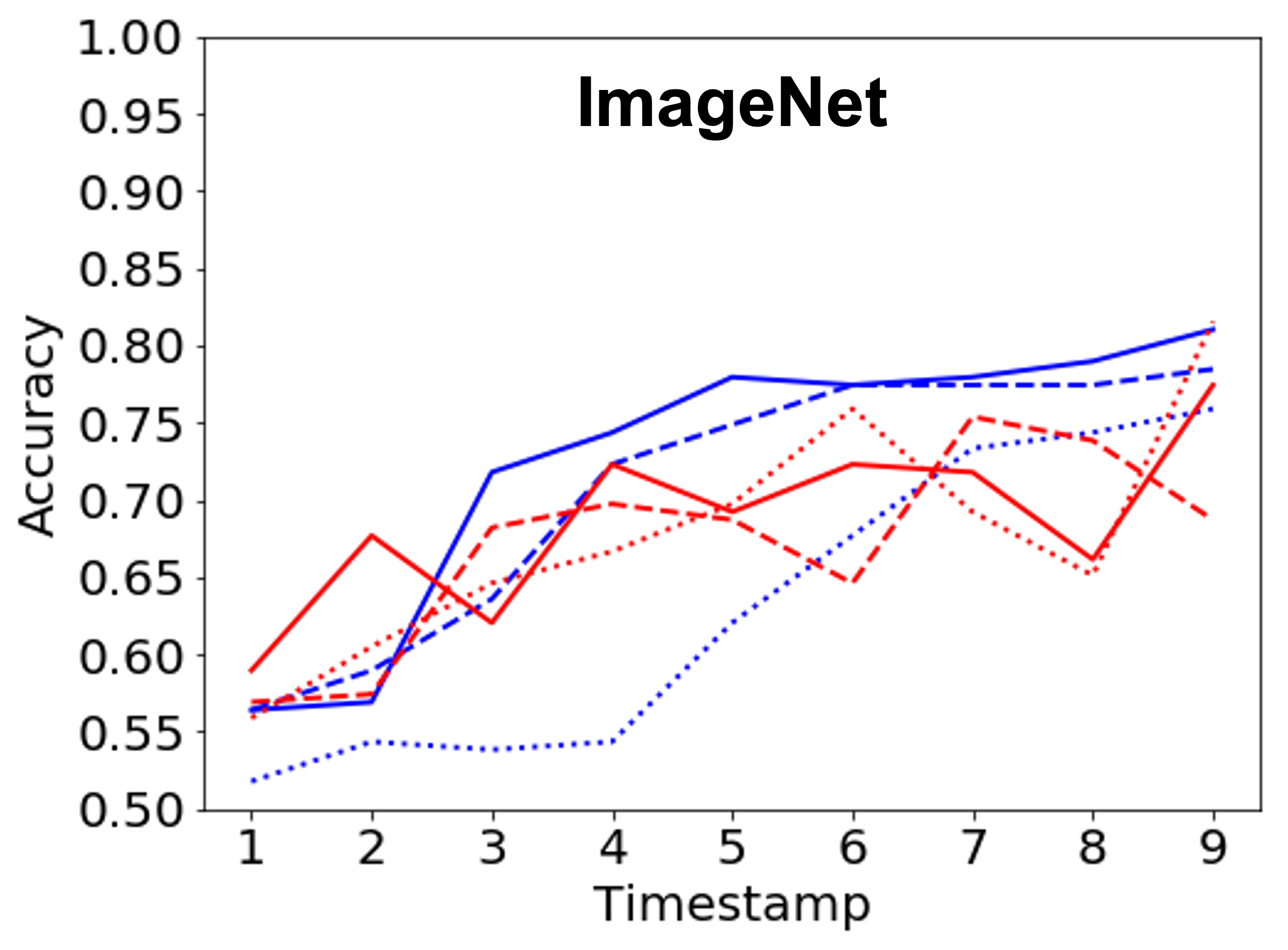}
      \caption{LeNet: deg}
  	\label{fig:img-len-degree}
  \end{subfigure}
  \hfill 
     \begin{subfigure}[t]{0.16\textwidth}
      \centering
      \includegraphics[width=\textwidth,trim={0 0 0 1cm}]{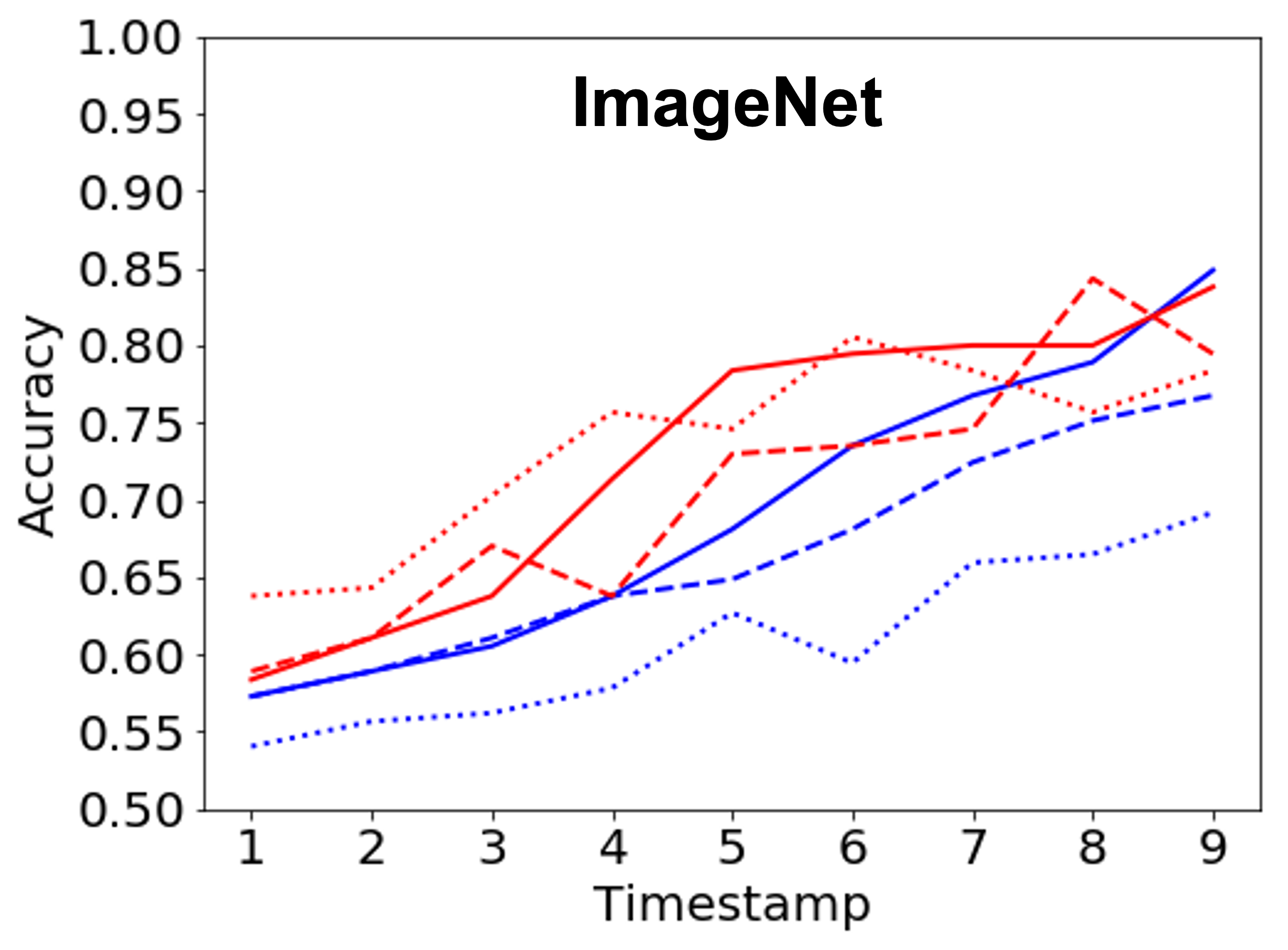}
      \caption{LeNet: evec}
      \label{fig:img-len-c}
  \end{subfigure}
    \hfill
   \begin{subfigure}[t]{0.16\textwidth}
      \centering
      \includegraphics[width=\textwidth,trim={0 0 0 0.1cm}]{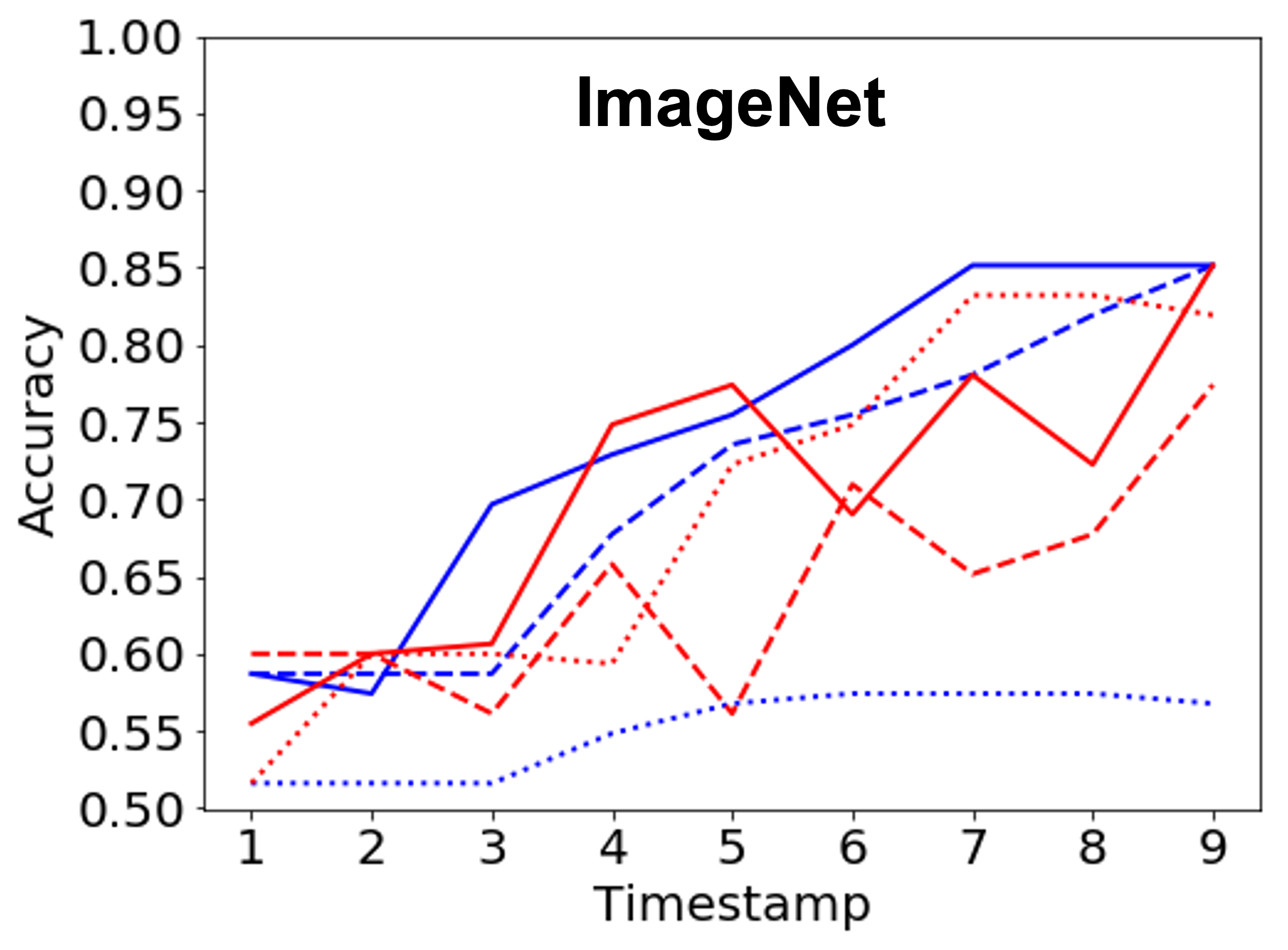}
      \caption{AlexNet: deg}
      \label{fig:img-al-deg}
  \end{subfigure}
  \hfill
   \begin{subfigure}[t]{0.16\textwidth}
      \centering
      \includegraphics[width=\textwidth,trim={0 0 0 0.1cm}]{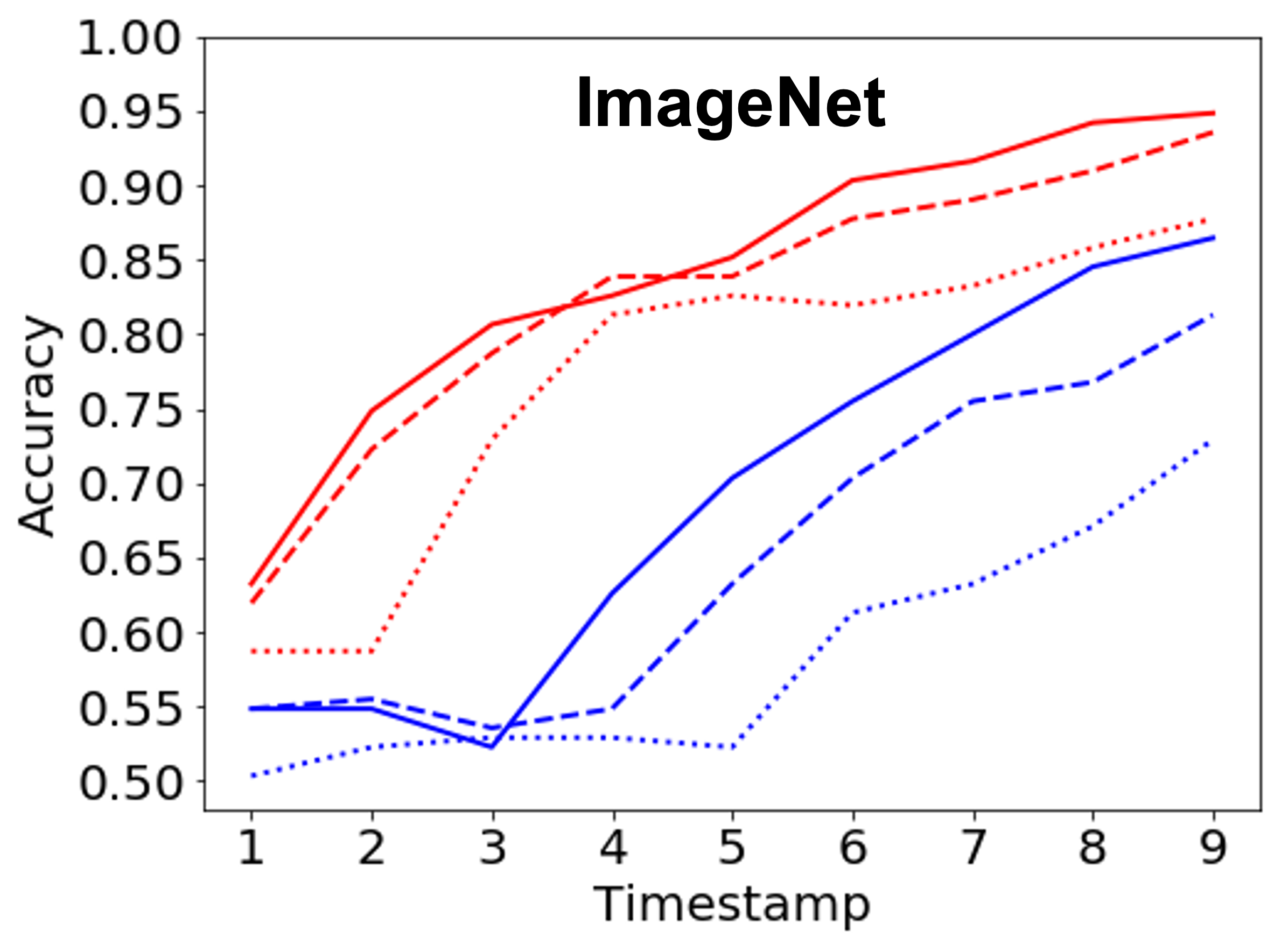}
      \caption{AlexNet: evec} 
      \label{fig:img-al-cen}
  \end{subfigure}
    \hfill
   \begin{subfigure}[t]{0.16\textwidth}
      \centering
      \includegraphics[width=\textwidth,trim={0 0 0 0.1cm}, clip]{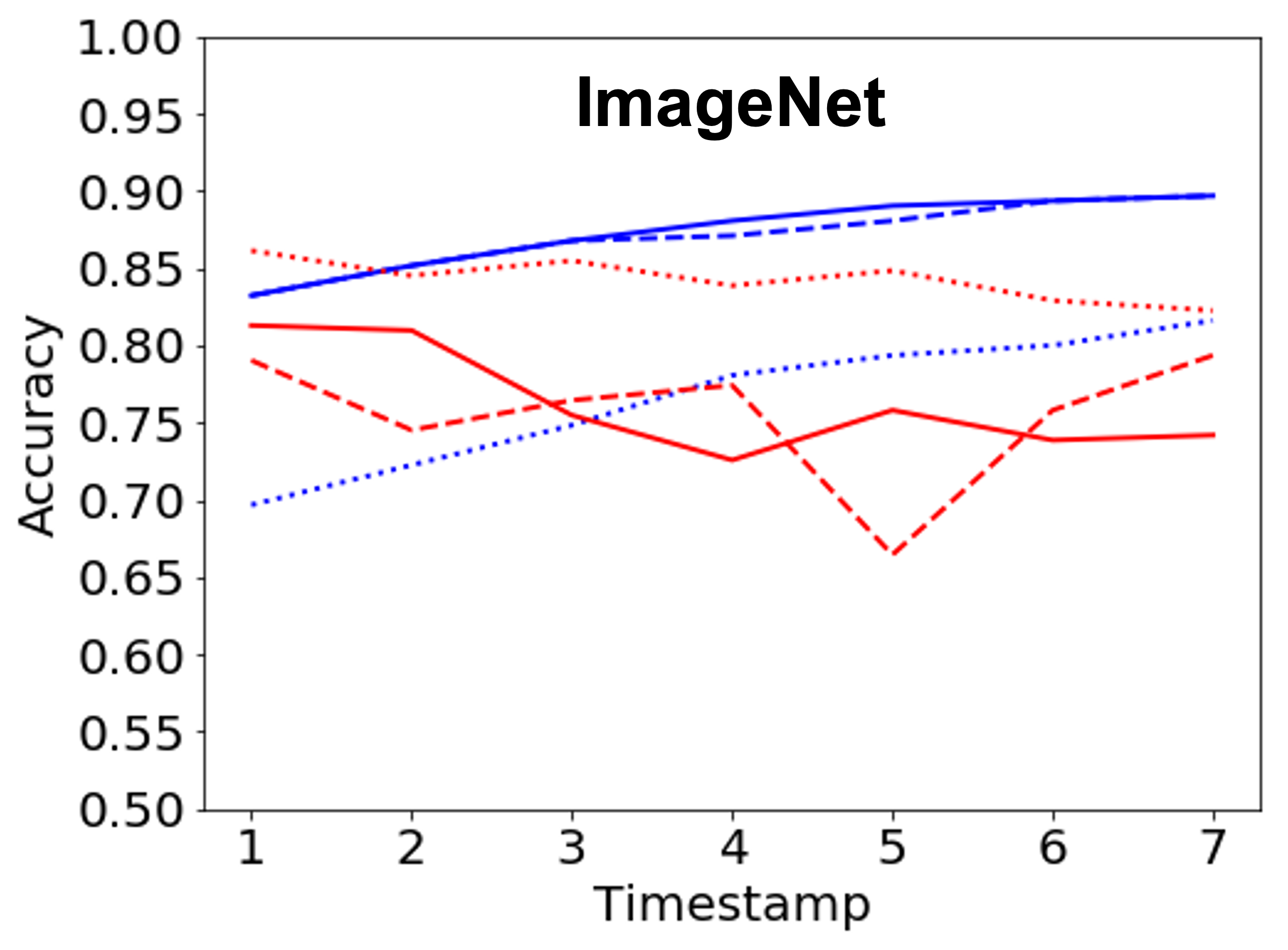}
      \caption{ResNet-50: deg }
      \label{fig:Res50-deg}
  \end{subfigure}
  \hfill
   \begin{subfigure}[t]{0.16\textwidth}
      \centering
      \includegraphics[width=\textwidth,trim={0 0 0 0.1cm}, clip]{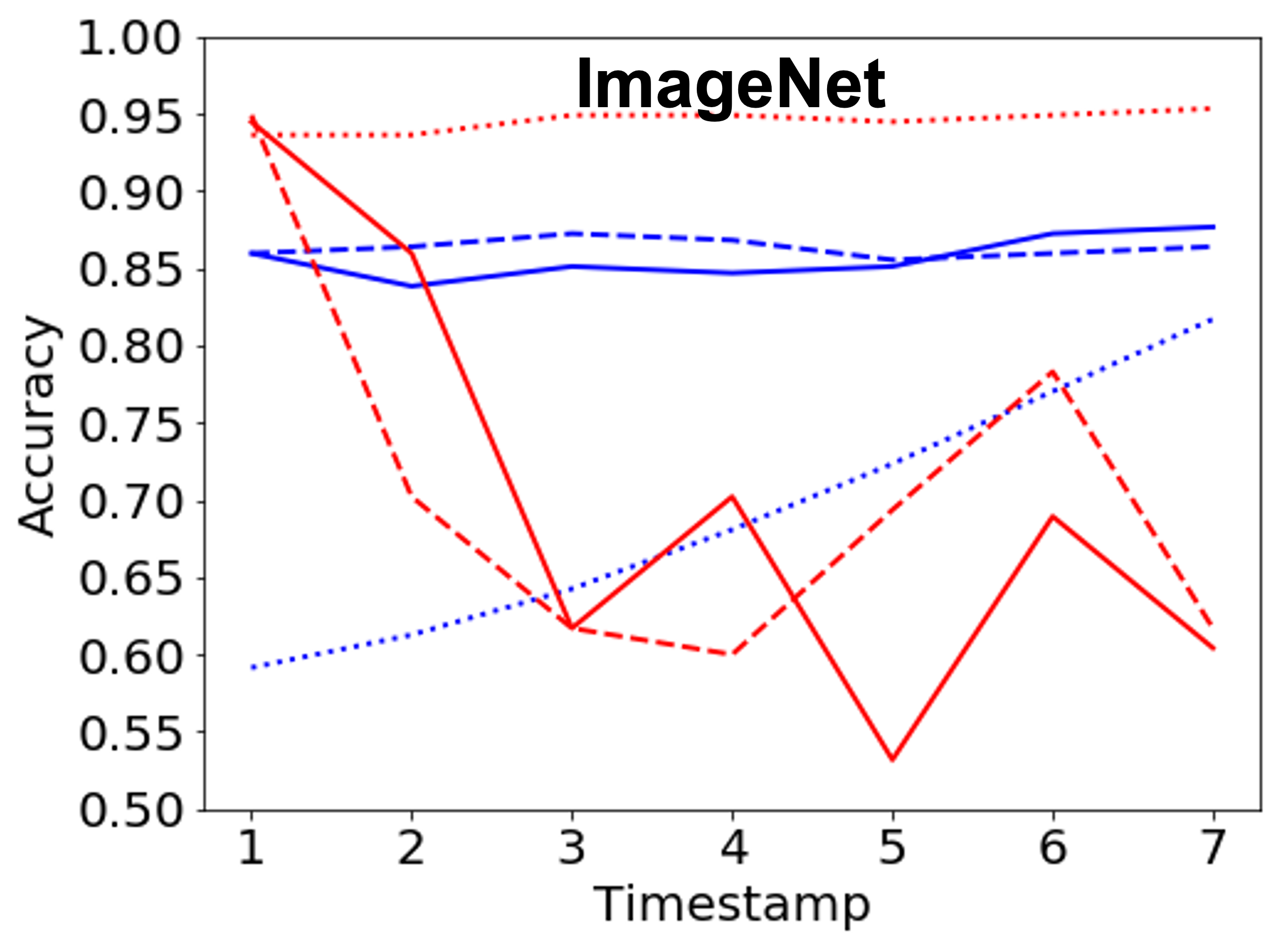}
      \caption{ResNet-50: evec }
      \label{fig:res50-cen}
  \end{subfigure}
  
    \caption{NN performance classification on CIFAR-10 (a-f) and ImageNet (g-l): `deg' for degree-based and `evec' for eigenvector centrality-based temporal signature. 
    For ImageNet, eigenvector-based signatures 
  yield higher performance compared to the degree-based ones.
  }
\label{fig:cifar-classification}
\end{figure*}

\vspace{0.13cm}
\noindent \textbf{(Q1) NN Performance Prediction  - Per Architecture.}
\textit{Task setup.}
We cast the NN performance prediction as a classification task where the generated temporal graphs are labeled as high and low accuracy based on the performance of their corresponding NNs. Table \ref{tab:NN} lists the threshold value chosen for low and high accuracy labels based on the final accuracy range of trained NNs, as well as the early stopping epochs for each architecture. Five-fold cross validation is used to predict the label of the test graphs using SVM and MLP, where the input is the set of temporal signatures 
$\{\vecf_{tr_1}^t, \vecf_{tr_2}^t, ...,\vecf_{tr_n}^t\}$. 
In Fig.~\ref{fig:cifar-classification}, per classifier we use the three signatures described in Step (S3), denoted as: no suffix, -wAvg, -expAvg.

\vspace{0.13cm}
\noindent \textit{Results.} We present the results for %
both types of 
signatures for the temporal graphs corresponding to the training dynamics of 
different architectures on 
CIFAR-10 and ImageNet in Fig.~\ref{fig:cifar-classification} (top and bottom, resp). 
In all the cases, classification accuracy of 80-95\% is achieved in less than 10 training epochs. For degree-based signatures, SVM tends to outperform MLP, while the trend is reversed for eigenvector-based signatures. 
For example, for both VGG and AlexNet for the CIFAR-10 image classification task, MLP can predict the performance with accuracy $\sim$95\% using the eigenvector-based signatures from the first 6 training epochs; the same trend is observed on ImageNet for the LeNet and AlexNet architectures.  
 In all the cases, both classifiers reach performance over 80\%-90\% significantly before the early stopping point for all the architectures.  

\vspace{0.05cm}
\noindent \textit{Comparison to baselines.} {Table \ref{tab:baseline} summarizes the classification accuracy of the last timestamp $t$ of our proposed framework (e.g., $t=9$ for LeNet) and the baselines. For all the architectures except for VGG, both signature types of our graph-aware framework achieve significantly higher accuracy than the three graph-agnostic baselines. For VGG (CIFAR-10), our degree-based graph signature outperforms the baselines, while the eigenvector-based signature achieves comparable performance to $B_{W^\prime}$, which replaces our graph features with the flattened weights in the last NN layer. 
The consistently high accuracy of our framework in
this task compared to the graph-agnostic baselines illustrates the utility of our graph representation and feature extraction, and the insufficiency of leveraging directly the learned weights.}
Finally, replacing our efficient  rolled graph representation with the baseline unrolled representation leads to significantly worse performance and longer runtime, which justifies the utility of our representation.  
Due to extensive memory and space requirements of the unrolled representation, we report results only  for LeNet (t=9) and AlexNet (t=14) on CIFAR-10. 

\begin{table}[t]
\scriptsize
\vspace{-0.1cm}
  \caption{Comparison of  %
  classification accuracy between baselines and our graph-based framework. {OOM: Out Of Memory.}}
  \vspace{-0.3cm}
  \label{tab:baseline}
  \centering
  \resizebox{.98\columnwidth}{!}{
  \setlength{\tabcolsep}{3pt}
  \begin{tabular}{lrrrrrcrr}
    \toprule
    \multicolumn{2}{c}{} & \multicolumn{3}{c}{\bf  CIFAR-10 } && \multicolumn{3}{c}{\bf ImageNet}                  \\
    \cmidrule(r){3-6} \cmidrule(r){8-9}
       && \textbf{LeNet} & \textbf{AlexNet} & \textbf{VGG} &  \textbf{ResNet-44} && \textbf{AlexNet} &  \textbf{ResNet-50}\\
    \midrule
    &$B_{W_l}$ &  0.63   & 0.61& 0.65& 0.48 &    &  0.64 & 0.53 \\
     &$B_{\hat{W}}$ &  0.8 & 0.74&0.51 & 0.74 &    & 0.66 & 0.56 \\
    \textbf{SVM} &$B_{W^\prime}$&  0.75 &0.82& 0.88   & 0.72 &    &  0.67 & 0.6  \\
     &\textbf{unroll-deg}  & 0.92  &0.49 & OOM &  OOM &   & OOM & OOM\\ \cmidrule{2-9}
    &\textbf{Ours-deg}  &   \textbf{0.99}&0.87 &\textbf{0.97} & \textbf{0.95} &    & 0.85  & \textbf{0.88}  \\
    &\textbf{Ours-evec}  &  0.81 & \textbf{0.94}&0.86 &0.88      &    & \textbf{0.86}  &  \textbf{0.88} \\
    \bottomrule
    
    &$B_{W_l}$ &  0.68   & 0.83& 0.67& 0.47 &    &  0.64 & 0.68 \\
     &$B_{\hat{W}}$ &  0.55 & 0.57&0.58 & 0.61 &    & 0.62 & 0.55 \\
    \textbf{MLP} &$B_{W^\prime}$&  0.58 &0.75& 0.52   & 0.66 &    &  0.67 & 0.58  \\
    
    &\textbf{unroll-deg}  & 0.93  &0.89 & OOM  &   OOM &   & OOM & OOM \\ \cmidrule{2-9}
    
    &\textbf{Ours-deg}  &   \textbf{0.98}&0.91 &\textbf{0.96} & 0.88 &    &  0.85 & \textbf{0.74}  \\
    & \textbf{Ours-evec}  &  0.91 & \textbf{0.96}&0.85 &\textbf{0.96}      &    & \textbf{0.94}  &  \textbf{0.70} \\
    \bottomrule
  \end{tabular}
   }
   \vspace{-0.15cm}
\end{table}

\vspace{0.13cm}
\noindent \textbf{(Q2) NN Performance Prediction  - Generalizing to Unseen Architectures.}
\label{app:generalization}
\textit{Task setup.} To show that our proposed graph representation and signatures are general, we fully train a small set of different NN architectures (and hyperparameters), and predict the performance on unseen NN architectures. Two sets of experiments were considered:
\textbf{(1)} Train our classifier on the smaller ResNet architectures  (ResNet-32 for CIFAR-10, and ResNet-34 for ImageNet), and test the performance of the larger ResNet architectures (ResNet-44 for CIFAR-10, and ResNet-50 for ImageNet);  
\textbf{(2)} Train our classifier on the combination of old architectures (VGG, AlexNet and LeNet) to predict the accuracy of a newer architecture (ResNet).

\begin{figure}[t]
\centering
\begin{subfigure}[t]{0.15\textwidth}
\includegraphics[width=\textwidth,trim={0 0 0 0.8cm}]{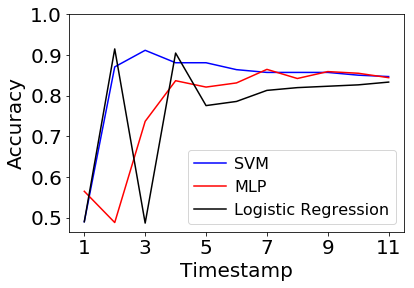}
\caption{CIFAR, ResNet44 
}
\label{tr-res32-res44}
\end{subfigure}
\begin{subfigure}[t]{0.15\textwidth}
\includegraphics[width=\textwidth,trim={0 0 0 0.8cm}]{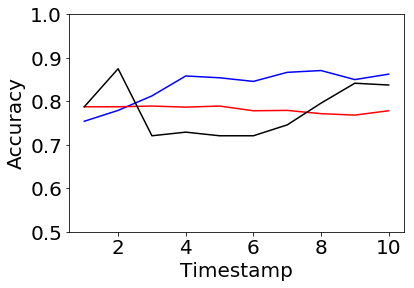}
\caption{ImgNet, ResNet50} 

\label{tr-res34-res50}
\end{subfigure}
\begin{subfigure}[t]{0.15\textwidth}
\includegraphics[width=\textwidth,trim={0 0 0 0.8cm}]{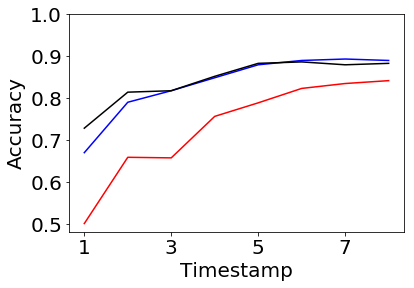}
\caption{CIFAR, ResNet32}
\label{tl-all-res32}
\end{subfigure}
\caption{NN classification to generalize \textbf{degree} signatures to unseen architectures. \textbf{(a)} Train: ResNet-32, test:ResNet-44. \textbf{(b)} Train: ResNet-34, test: ResNet-50. \textbf{(c)} Train: LeNet, AlexNet, VGG, test:  ReNet-32.
}
\label{fig:tl}
\end{figure}

\vspace{0.1cm}
\noindent \textit{Results.} 
In Figs.~\ref{tr-res32-res44} and \ref{tr-res34-res50} (exp. 1), we observe that our proposed framework is able to accurately predict the performance level of the previously unseen, large ResNet architectures. Our results show that the proposed temporal signatures can be used in a generalized scenario to predict the accuracy level of the same architecture with different numbers of layers (on the same dataset). This generalization from small to bigger architectures is important since it is faster to train the smaller architectures. 
In Fig. \ref{tl-all-res32} (exp. 2), we see that our proposed method also successfully predicts the performance level of a new architecture (i.e. ResNet) when the training set is a combination of older architectures (LeNet, VGG, AlexNet).

\begin{figure}[t]
    \centering
    \begin{subfigure}[t]{0.15\textwidth}
    \includegraphics[width=\textwidth,trim={0 0 0 1cm}]{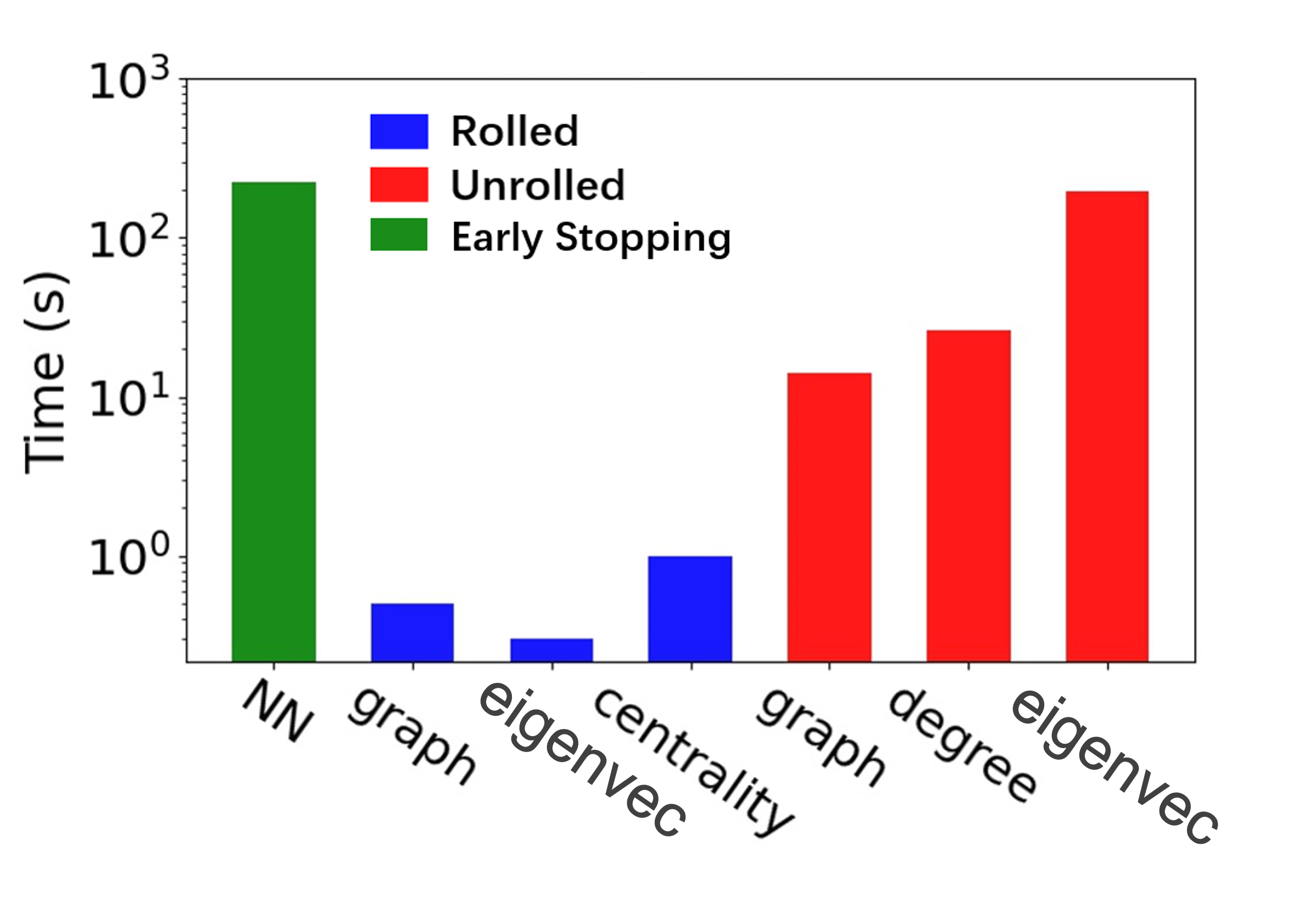}
    \label{time-len}
    \vspace{-0.6cm}
    \caption{LeNet}
    \end{subfigure}
    \hfill
    \begin{subfigure}[t]{0.15\textwidth}
    \includegraphics[width=\textwidth,trim={0 0 0 1cm}]{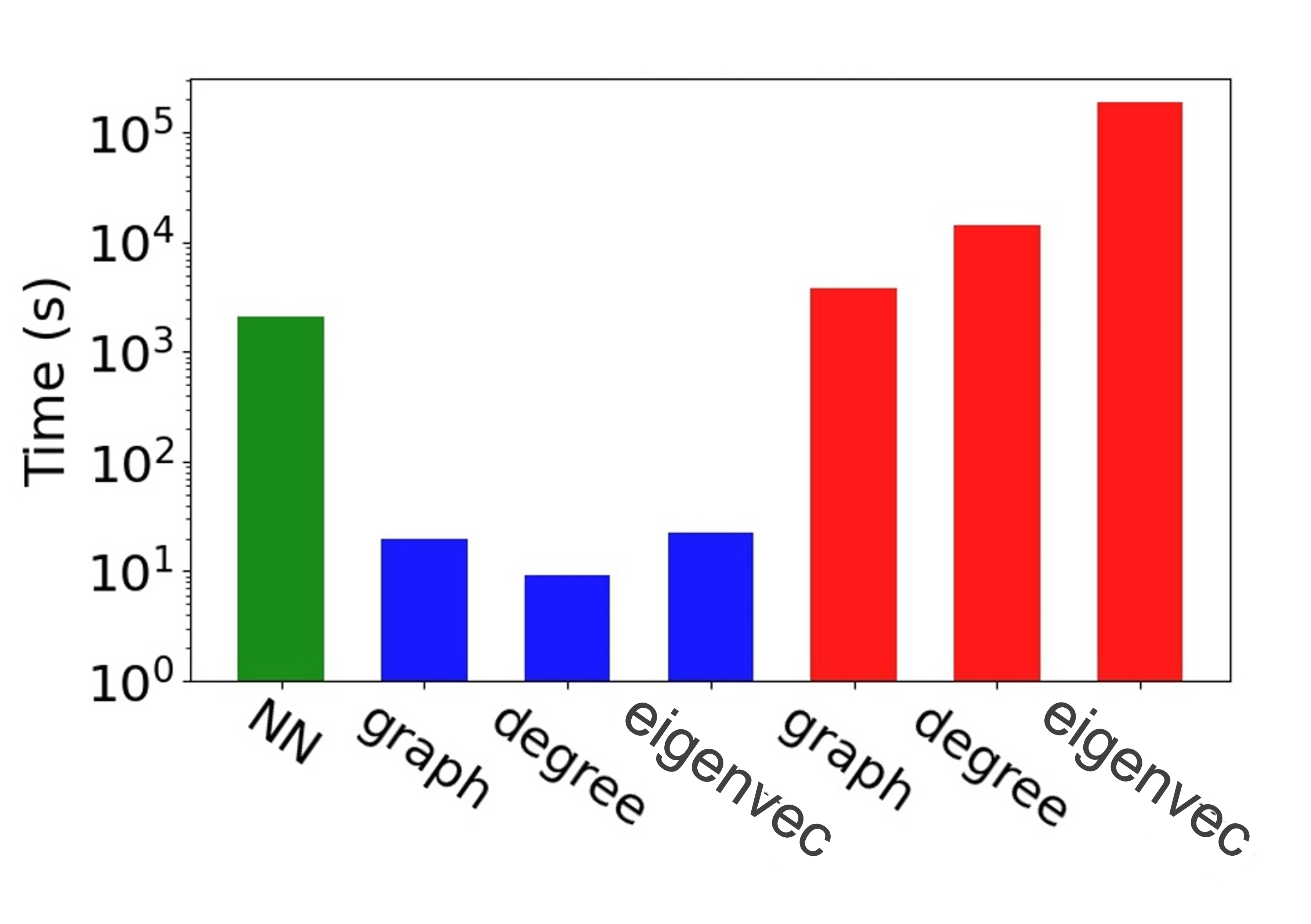}
    \label{time-vgg}
    \vspace{-0.6cm}
    \caption{VGG}
    \end{subfigure}
    \hfill
    \begin{subfigure}[t]{0.15\textwidth}
    \includegraphics[width=\textwidth,trim={0 0 0 0.8cm}]{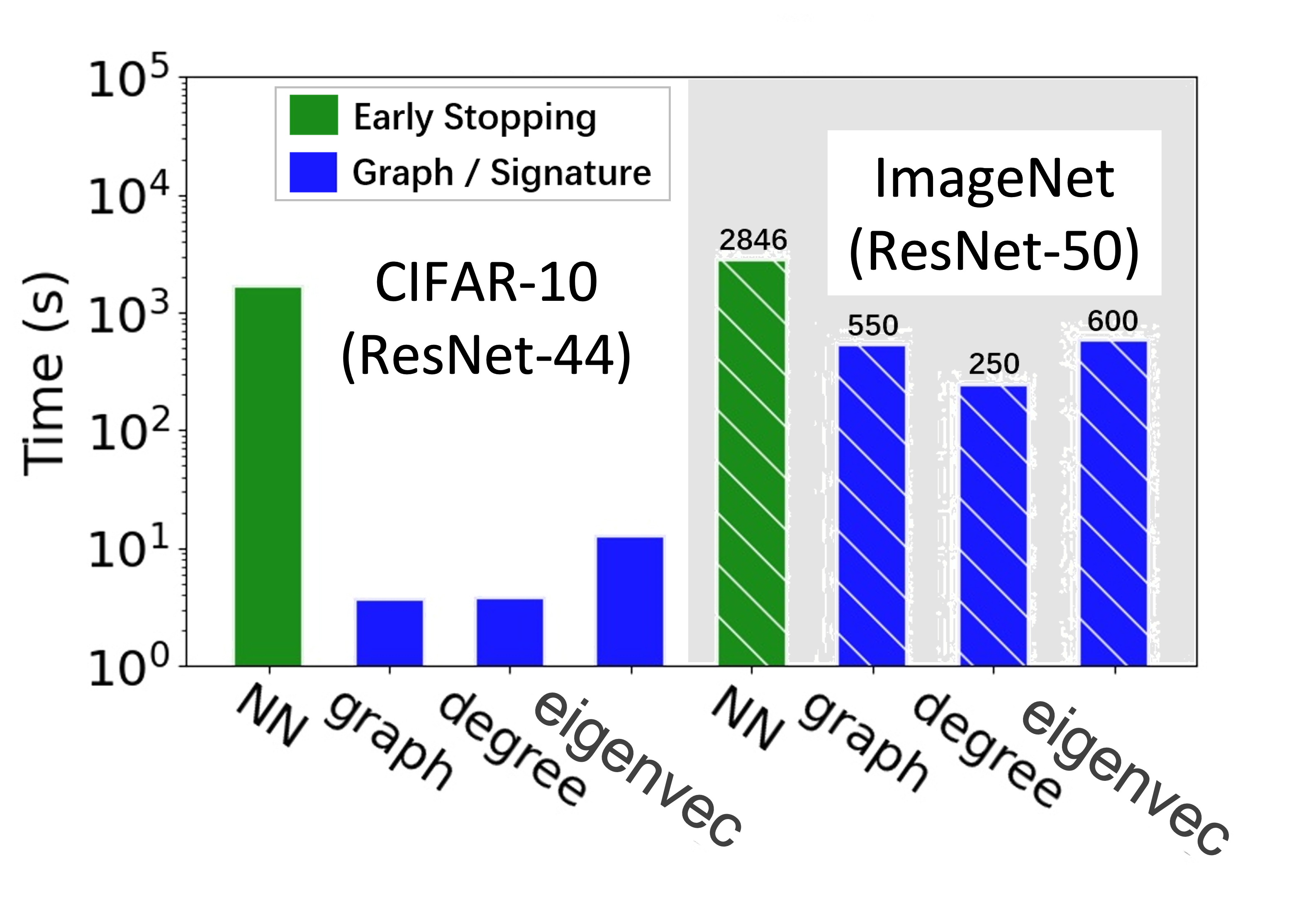}
    \label{time-res}
    \vspace{-0.6cm}
    \caption{ResNet}
    \end{subfigure}
    \caption{Avg runtime for the NN training, the graph generation (\textbf{S1}) and feature extraction (\textbf{S2}) on: (\textbf{a}) CIFAR-10 with LeNet; (\textbf{b}) CIFAR-10 with VGG; and (\textbf{c}) CIFAR-10 (ResNet-44) and ImageNet (ResNet-50).
    }
\label{fig:lenet-time}
\end{figure}

\vspace{0.13cm}
\noindent\textbf{(Q3) Time efficiency and early stopping.} 
Figure~\ref{fig:lenet-time} depicts the total average runtime of NN training for early stopping of each architecture (green bars) in comparison to the average runtime of graph generation, degree calculation and eigenvector centrality calculation for the number of epochs  needed for that architecture to achieve the highest accuracy in classification task. For all  the architectures, our proposed rolled graph representation (in blue) can achieve a high-accuracy prediction much faster than the early stopping approach of NN training. But for the baseline unrolled representation (in red), this is only true for LeNet. As the size of NN increases, the  unrolled graph generation gets slower and the corresponding graph-based approach is slower than early stopping. 

\vspace{0.15cm}
\vspace{0.15cm}

%% file: PAGES_v3/050conclusion.tex
We investigated the early training dynamics of NNs from a time-evolving graph perspective. 
To the best of our knowledge, we are the first to model the NN training dynamics and structure as a temporal graph. 
We coupled this representation with a new, compact graph model for convolutional layers. 
Then, we showed that a simple, temporal graph signature based on summary statistics of the degree or eigenvector centrality distributions over \textit{only a few epochs} 
can be used as a strong predictor variable to estimate the accuracy of NNs in downstream tasks. 
Exploring the role of our efficient proposed framework for early stopping is a promising future direction.